\documentclass[journal]{IEEEtran}
\usepackage{cite}
\usepackage{amsmath,amssymb,amsfonts}
\usepackage{algorithm,algorithmic}
\usepackage{graphicx}
\usepackage{textcomp}
\usepackage{xcolor}
\usepackage{amsfonts}
\usepackage{multirow}
\usepackage{booktabs} 
\usepackage{diagbox}
\usepackage{tikz}
\usepackage{pgfplots}
\pgfplotsset{width=0.5\columnwidth,compat=1.18}
\pgfplotsset{every axis/.append style={
                    label style={font=\small},
                    }}
\usepackage{hyperref}
\hypersetup{colorlinks=true,
            linkcolor=blue,
            anchorcolor=blue,
            citecolor=blue,
            urlcolor=black,
            }

\begin{document}
\title{Rethinking Barely-Supervised Volumetric Medical Image Segmentation from an Unsupervised Domain Adaptation Perspective}
\author{Zhiqiang Shen, Peng Cao, Junming Su, Jinzhu Yang, and Osmar R. Zaiane
\thanks{This research was supported by the National Natural Science Foundation of China (No.62076059), the Science and Technology Joint Project of Liaoning Province (2023JH2/101700367) and the Fundamental Research Funds for the Central Universities (No. ZX20240193). Corresponding Author: Peng Cao.}
\thanks{Zhiqiang Shen, Peng Cao, Junming Su, and Jinzhu Yang are with the School of Computer Science and Engineering, Northeastern University, Shenyang 110819, China, and also with the Key Laboratory of Intelligent Computing in Medical Image of Ministry of Education, Northeastern University, Shenyang 110819, China (e-mail: xxszqyy@gmail.com; caopeng@cse.neu.edu.cn).}
\thanks{Osmar R. Zaiane is with the Alberta Machine Intelligence Institute, University of Alberta, Edmonton, Canada.}
}

\maketitle

\begin{abstract}
This paper investigates an extremely challenging problem: barely-supervised volumetric medical image segmentation (BSS). A BSS training dataset consists of two parts: 1) a barely-annotated labeled set, where each labeled image contains only a single-slice annotation, and 2) an unlabeled set comprising numerous unlabeled volumetric images.
State-of-the-art BSS methods employ a registration-based paradigm, which uses inter-slice image registration to propagate single-slice annotations into volumetric pseudo labels, constructing a completely annotated labeled set, to which a semi-supervised segmentation scheme can be applied.
However, the paradigm has a critical limitation: the pseudo-labels generated by image registration are unreliable and noisy.
Motivated by this, we propose a new perspective: instead of solving BSS within a semi-supervised learning scheme, this work formulates BSS as an unsupervised domain adaptation problem.
To this end, we propose a novel BSS framework, \textbf{B}arely-supervised learning \textbf{via} unsupervised domain \textbf{A}daptation (BvA), as an alternative to the dominant registration paradigm.
Specifically, we first design a novel noise-free labeled data construction algorithm (NFC) for slice-to-volume labeled data synthesis. Then, we introduce a frequency and spatial Mix-Up strategy (FSX) to mitigate the domain shifts.
Extensive experiments demonstrate that our method provides a promising alternative for BSS. Remarkably, the proposed method, trained on the left atrial segmentation dataset with \textbf{only one} barely-labeled image, achieves a Dice score of 81.20\%, outperforming the state-of-the-art by 61.71\%.
The code is available at \href{https://github.com/Senyh/BvA}{\textit{\texttt{https://github.com/Senyh/BvA}}}.
\end{abstract}

\begin{IEEEkeywords}
Barely-Supervised Learning, Medical Image Segmentation, Semi-Supervised Learning, Unsupervised Domain Adaptation
\end{IEEEkeywords}

\section{Introduction}
\label{sec:intro}
\begin{figure}[!t]
\centering
\includegraphics[width=\columnwidth]{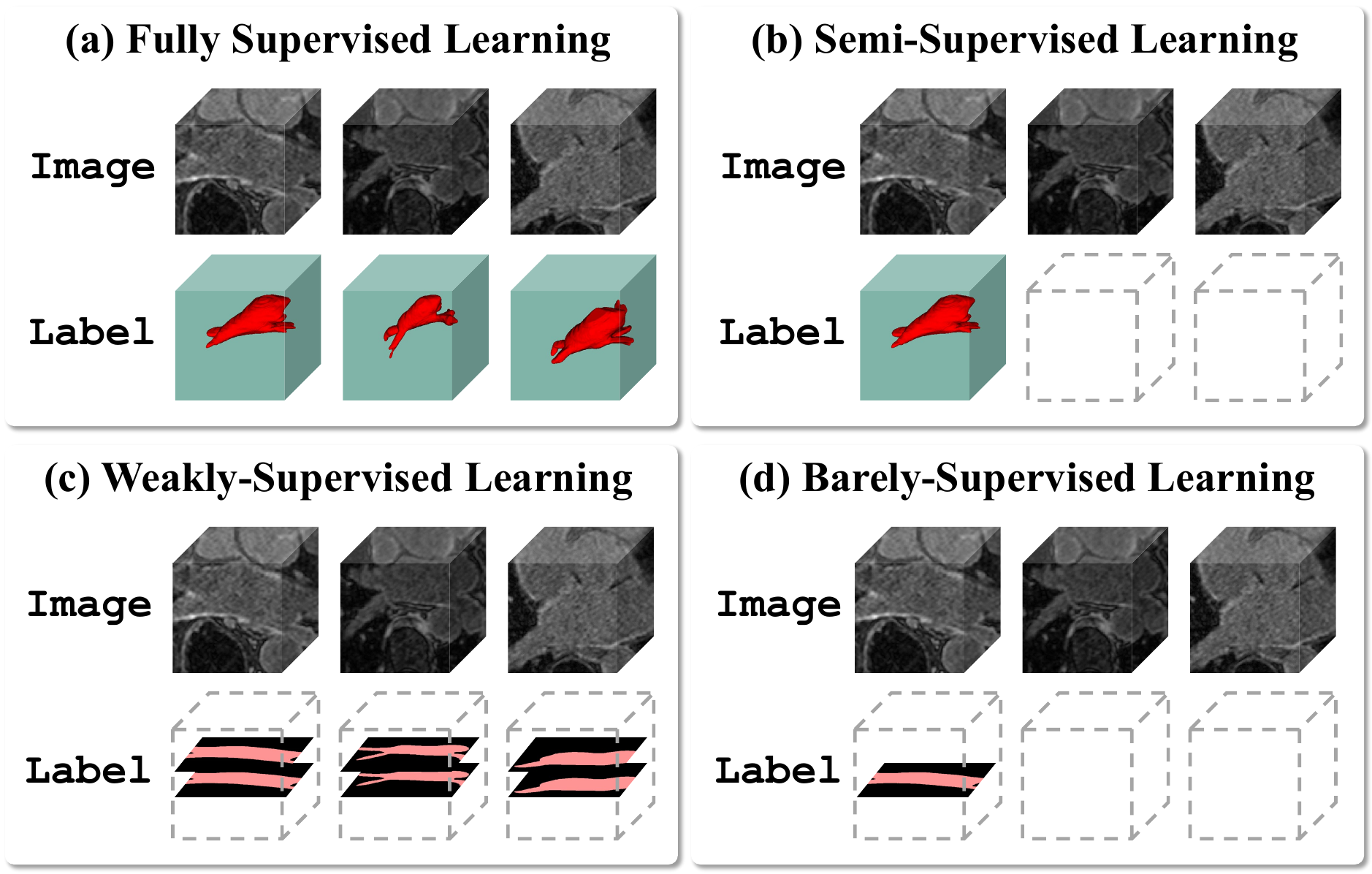}
\caption{Illustration of different learning paradigms. (a) fully supervised learning, (b) semi-supervised learning, (c) weakly-supervised learning, and (d) barely-supervised learning.} 
\label{fig:bva_intro_paradigms}
\end{figure}

\IEEEPARstart{M}{edical} image segmentation is essential for computer-aided diagnosis, providing accurate localization and delineation of organs and tumors for disease progression monitoring and surgical planning. 
Considerable advances have been made based on fully supervised learning (FSL), which relies on large-scale \textbf{fully and completely} annotated datasets that the entire dataset is fully annotated and each sample has a complete label [Fig. \ref{fig:bva_intro_paradigms}(a)]. 
However, annotating medical images, especially volumetric images with hundreds of slices, at the pixel level is laborious and requires expert knowledge, resulting in a significant annotation burden. 
Semi-supervised learning (SSL) \cite{yu2019uncertainty,shen2023co,yang2023revisiting} and weakly-supervised learning (WSL) \cite{zhang2022shapepu,zhang2022cyclemix} are two prevailing schemes for alleviating the annotation burden on medical image segmentation. 
SSL learns from a dataset that is \textbf{partially but completely} annotated, consisting of a small number of labeled images with complete annotations and a large number of unlabeled images [Fig. \ref{fig:bva_intro_paradigms}(b)]; WSL requires a \textbf{fully but incompletely} annotated dataset, where images have incomplete annotations, such as bounding boxes, scribbles, or only a few slices per volumetric image, as shown in Fig. \ref{fig:bva_intro_paradigms}(c).
Nevertheless, hundreds and thousands of slices still need to be labeled at the pixel level in volumetric medical image segmentation tasks.

%---------------------------------------------------------------
%---------------------------------------------------------------

%---------------------------------------------------------------
\begin{figure*}[!t]
\centering
\includegraphics[width=\textwidth]{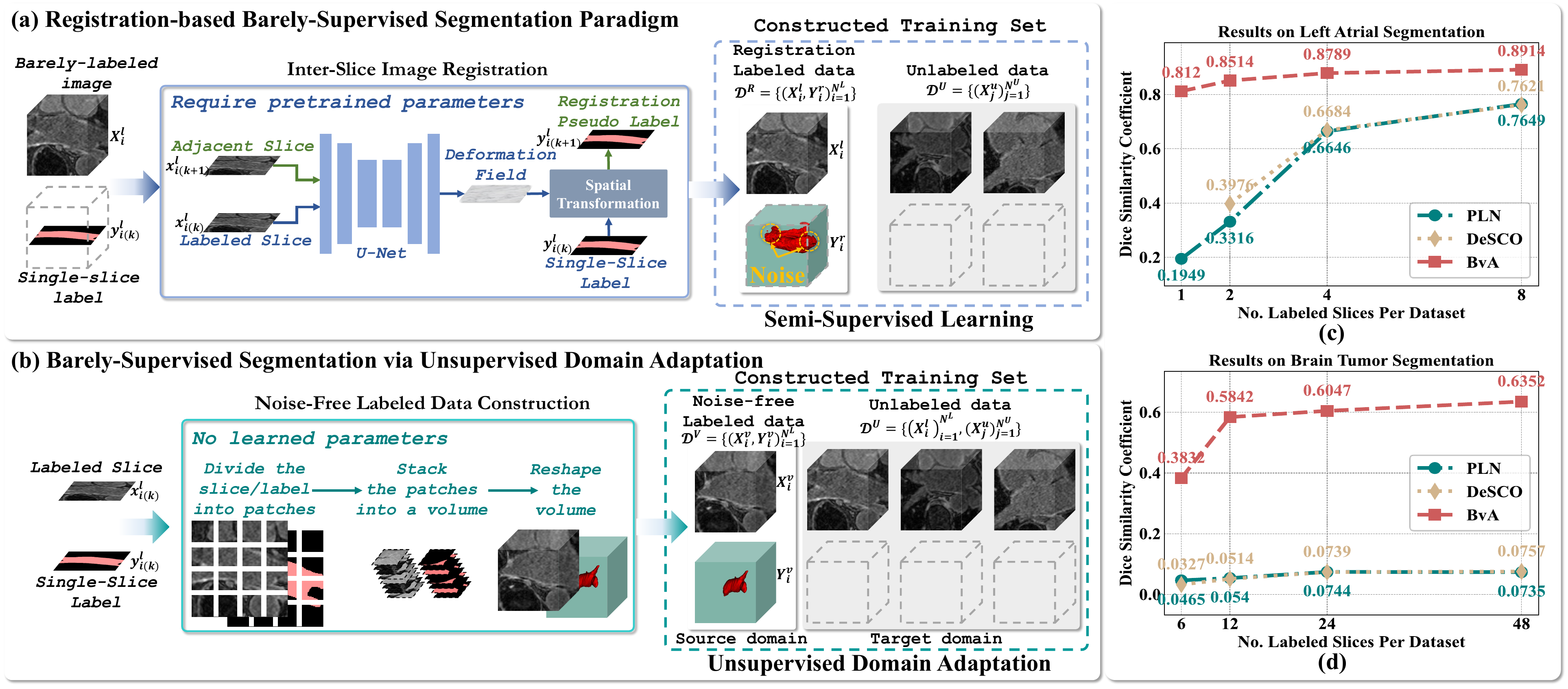}
\caption{Illustration of (a) the registration-based paradigm, (b) our BvA framework, and (c-d) barely-supervised segmentation results on the left atrial and brain tumor segmentation tasks. Note that DeSCO requires two orthogonal labeled slices per image; therefore, it is unavailable in situations where the training set contains only one labeled slice.} 
\label{fig:bva_intro_frameworks}
\end{figure*}
%---------------------------------------------------------------

Barely-supervised learning (BSL) based volumetric medical image segmentation, abbreviated as BSS, has the potential to reduce annotation costs further \cite{bitarafan20203d}, with the setting [Fig. \ref{fig:bva_intro_paradigms}(d)] only requiring a \textbf{partially and incompletely} annotated dataset, which comprises a barely-annotated labeled set with single-slice annotations $\mathcal{D}^L = \{(X^l_i, y^l_{i(k)})_{i=1}^{N_L}\}$ and an unlabeled set $\mathcal{D}^U = \{(X^u_j)_{j=1}^{N^U}\}$.
The key challenge in BSS is how to generate volumetric labels for barely-annotated images and unlabeled images.
State-of-the-art BSS methods \cite{li2022pln,cai2023orthogonal} build upon a registration-based paradigm, which equips with an inter-slice image registration module.
As illustrated in Fig.~\ref{fig:bva_intro_frameworks}(a), 
for each barely-annotated volumetric image $X^l_i$ with a single-slice annotation $y^l_{i(k)}$, the inter-slice registration module gradually propagates labels between adjacent slices to predict a registration pseudo label $Y^r_i$, transforming the barely-annotated labeled set $\mathcal{D}^L$ into a completely annotated labeled set $\mathcal{D}^R = \{(X^l_i, Y^r_{i})_{i=1}^{N_L}\}$. 
Then, the constructed labeled set $\mathcal{D}^R$ is combined with the original unlabeled set $\mathcal{D}^U$ to form a new training set, on which a semi-supervised learning procedure is conducted.
However, the paradigm has a critical limitation: the pseudo labels generated by 2D registration are unreliable and noisy, degrading the extraction of accurate supervisory signals from barely annotated images.
We conducted a pilot experiment to investigate the effect of this limitation on BSS. 
As shown in Fig. \ref{fig:bva_intro_frameworks}(c), PLN \cite{li2022pln} and DeSCO \cite{cai2023orthogonal} obtain inferior performance on left atrial segmentation, especially in scenarios with only one or two annotated slices (per dataset) \footnote{This means that the entire dataset has only one or two barely-annotated images that were labeled with single-slice annotations.} in the training set. 
The results become even more unsatisfactory in the more challenging brain tumor segmentation task with heterogeneous tumors [Fig. \ref{fig:bva_intro_frameworks}(c)]. 
Due to the extreme scarcity of labeled data, reliable volumetric pseudo-labels cannot be generated from original single-slice annotations. Therefore, this work pinpoints the key problem: \textit{how to excavate volumetric supervisory information from reliable single-slice annotations to train the segmentation model without relying on generating slice-wise registration pseudo-labels.}

To this end, we propose a novel BSS framework, \textbf{B}arely-supervised learning \textbf{via} unsupervised domain \textbf{A}daptation (BvA) [Fig. \ref{fig:bva_intro_frameworks}(b)].
One can observe from Fig. \ref{fig:bva_intro_frameworks}(c-d) that, BvA consistently outperforms the registration-based methods by a large margin in terms of the Dice Similarity Coefficient (DSC), especially in the case where the entire training set contains \textbf{only one} labeled slice.
For example, BvA surpasses PLN \cite{li2022pln} with a DSC score of 61.71\% in the case of \textbf{only one} labeled slice per dataset. 
Conceptually, instead of solving BSS using the registration-based paradigm, where registration pseudo-labels are often noisy and unreliable, we formulate BSS as an unsupervised domain adaptation (UDA) problem.
An intuitive solution is to leverage a UDA scheme based on the registration paradigm to address the BSS problem.
However, this approach is infeasible as UDA requires reliable annotations for source domain images, while the pseudo-labels generated by image registration are, as mentioned above, unreliable and noisy.
Instead, we introduce a noise-free labeled data construction algorithm (NFC) for slice-to-volume labeled data synthesis by deconstructing a single-annotated slice/label $x^l_{i(k)}/y^l_{i(k)}$ into patches and reconstructing a volumetric image $X^v_{i}/Y^v_{i}$ from the patches.
The idea of NFC lies in that the inter-patch similarity in a slice is akin to the inter-slice continuity of a volume.
Since the statistics of a single slice cannot represent those of the corresponding original image, there may be domain shifts between synthesized images and original images.
To mitigate domain shifts, we assume that a well-generalized model should behave smoothly across both source and target domains under small perturbations. Therefore, we propose a Frequency and Spatial Mix-Up strategy (FSX), which performs image Mix-Up \cite{zhang2018mixup} in the frequency \cite{yang2020fda,xu2021fourier} and spatial \cite{yun2019cutmix} domains to alleviate style and content shifts, respectively, in addressing the UDA problem [source domain: synthesized images $\mathcal{D}^V = \{(X^v_i, Y^v_{i})_{i=1}^{N_L}\}$, target domain: original images $\mathcal{D}^U = \{(X^l_i)_{i=1}^{N_L}, (X^u_j)_{j=1}^{N^U}\}$]. 
Note that we incorporate labeled images into the unlabeled image set to fully utilize the training data. Consequently, NFC synthetic images are used as the source domain data, while the original images (including both labeled and unlabeled images) serve as the target domain data.
Extensive experiments show that BvA significantly improves the state-of-the-art results on the left atrial and brain tumor segmentation benchmarks under both \textbf{barely-supervised segmentation} and \textbf{semi-supervised segmentation} settings. For example, BvA achieves 87.40\% in terms of DSC on the LA dataset with 5\% barely-labeled data (only 4 labeled slices in the training set) and a DSC of 58.42\% on the BraTS dataset with 5\% barely-labeled data (only 12 labeled slices), outperforming
PLN \cite{li2022pln} by 20.94\% and 53.01\%, respectively.

Our contributions mainly include:
\begin{itemize}
\item \textbf{New problem formulation}: To the best of our knowledge, this is the first work to formulate BSS from a UDA perspective, offering an alternative to the dominant registration paradigm in addressing BSS.
\item \textbf{New method}: We propose Noise-Free Labeled Data Construction (NFC) for constructing volumetric image-label pairs without requiring image registration. We further introduce a novel smoothness assumption for UDA. Based on this assumption, we design a Frequency and Spatial Mix-Up module (FSX) to mitigate the domain shifts between the synthesized and original images.
\item \textbf{Significant Performance Improvement}: Our method outperforms the state-of-the-art BSS approaches by average performance gains of about 20\% and 50\% in terms of DSC on the LA and BraTS datasets, respectively. Additionally, we found that for volumetric medical image segmentation tasks, annotating multiple images with single-slice annotations is a more effective sparse labeling strategy than annotating a single image with multi-slice annotations.
\end{itemize}

\section{Related work}
In the following section, we review related work on semi-supervised learning, weakly-supervised learning, barely-supervised learning, and unsupervised domain adaptation in the field of medical image segmentation.

\subsection{Semi-Supervised Learning}
Semi-supervised learning for medical image segmentation trains models from a partially but completely annotated dataset that consists of limited completely annotated labeled data and an arbitrary number of unlabeled images \cite{yu2019uncertainty,luo2021semi,luo2021urpc,wu2021semi,wu2022mutual,shen2023co,wang2023mcf,bai2023bidirectional}.
These studies, following the state-of-the-art technique of consistency regularization and pseudo-labeling \cite{sohn2020fixmatch,tarvainen2017mean,chen2021semi}, can be roughly divided into three branches: self-training-based \cite{luo2021urpc}, mean-teacher-based \cite{yu2019uncertainty,bai2023bidirectional}, and co-training-based \cite{luo2021semi,wu2021semi,wu2022mutual,shen2023co,wang2023mcf} approaches.
However, SSL methods cannot handle barely-supervised medical image segmentation tasks. In this paper, we take a step further to address the most challenging barely-supervised segmentation problem and propose a novel method for barely-supervised medical volumetric image segmentation.

\subsection{Weakly-Supervised Learning}
Weakly-supervised learning, which requires a fully but incompletely annotated dataset with incomplete annotations for each image, can be categorized according to the type of annotations, such as scribbles \cite{zhang2022shapepu,zhang2022cyclemix}, bounding boxes \cite{zhao2018deep,wei2023weakpolyp}, points \cite{qu2019weakly,qu2020weakly,lin2023nuclei}, as well as annotating a few slices \cite{bitarafan20203d}.
Most WSL methods exploit a weighted combination loss that includes a supervised term for sparse annotated data and a regularization term for unlabeled data.
However, since sparse annotations lack detailed shape information, WSL models struggle with delineating complex anatomical structures. This issue becomes even more severe in BSL scenarios.
To overcome this limitation, this paper explores a more challenging task of barely-supervised volumetric medical image segmentation, in which the limited labeled data only contain single-slice annotations.

\subsection{Barely-Supervised Learning}
Barely-supervised learning (BSL) was initially proposed in image recognition to address the issue of extremely limited supervision \cite{lucas2022barely}.
In the field of volumetric medical image segmentation, BSL aims to train models under a partially and incompletely annotated dataset involving only a limited number of single-slice labeled data and numerous unlabeled images \cite{bitarafan20203d}. 
State-of-the-art BSS methods generally employ a registration-based framework to reconstruct complete volumetric annotations from single-slice annotations, aiming to transform the barely-supervised learning problem into a semi-supervised learning problem \cite{bitarafan20203d,li2022pln,cai2023orthogonal}.
Since the registration-constructed volumetric labels are noisy, the registration procedure de facto results in a semi-supervised learning problem with extremely unreliable pseudo labels. In contrast, this study, from a novel perspective, proposes BvA to construct volumetric images with reliable labels using single-slice annotations, transforming the barely-supervised learning problem into an unsupervised domain adaptation problem.

\subsection{Unsupervised Domain Adaptation}
Unsupervised domain adaptation (UDA) aims to mitigate domain shifts between the source and target domains, assuming the availability of labeled data from the source domain and unlabeled data from the target domain.
Recently, numerous UDA methods have been proposed to bridge domain gaps, such as explicit minimization of distribution distance \cite{tzeng2014deep,wu2020cf}, implicit alignment via adversarial learning \cite{chen2019synergistic,chen2020unsupervised,pei2021disentangle,zhao2022uda}, and various data augmentation strategies to synthesize new image domains \cite{yang2020fda,huang2021fsdr}. In this work, we assume that a well-generalized model should behave smoothly across both source and target domains under small perturbations. Under this assumption, we leverage the image Mix-Up operation \cite{zhang2018mixup} in both spatial and frequency domains to mitigate the domain shifts between synthesized volumetric images and original images.

%---------------------------------------------------------------
\begin{figure*}[!t]
\includegraphics[width=\textwidth]{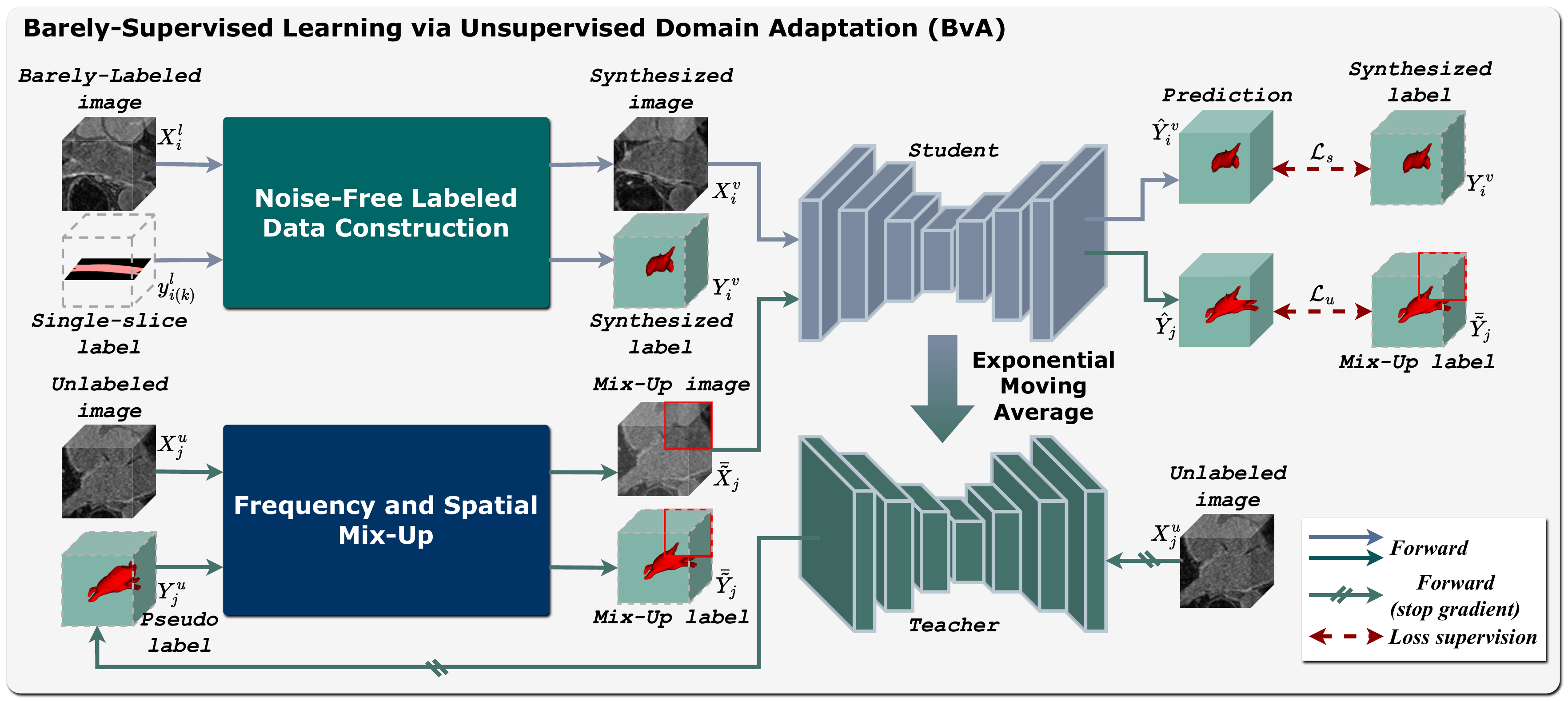}
\caption{Overview of the proposed \textbf{B}arely-supervised learning \textbf{via} unsupervised domain \textbf{A}daptation (BvA). BvA consists of 1) a noise-free labeled data construction algorithm (NFC) for generating volumetric labeled data, and 2) a frequency and spatial Mix-Up strategy (FSX) for alleviating domain shifts between the synthesized images and the original images. 
Note that BvA only requires the student model in the testing stage.
} 
\label{fig:bva_framework}
\end{figure*}
%---------------------------------------------------------------

%---------------------------------------------------------------
\definecolor{commentcolor}{RGB}{63,136,196}  
\newcommand{\PyComment}[1]{\ttfamily\textcolor{commentcolor}{\# #1}}  
\newcommand{\PyCode}[1]{\ttfamily\textcolor{black}{#1}} 

\begin{algorithm}[!t]
    \caption{BvA}
    \label{alg:BvA}
    \textbf{Input}: {$\mathcal{D}^L = \{(X^l_i, y^l_{i(k)})_{i=1}^{N_L}\}$ and $\mathcal{D}^U = \{(X^u_i)_{i=1}^{N^U}\}$}\\
    \textbf{Parameter}: $\theta$, $\theta^\prime$\\
    \textbf{Output}: $f(\cdot; \theta)$
    \begin{algorithmic}[1] 
            \FOR{each iteration}
                \STATE {\PyComment{Synthesize volumetric image-label pairs $(X^v_i, Y^v_i)$ using single-annotated slices $(x^l_{i(k)}, y^l_{i(k)})$}}
                \STATE {Divide the slice into patches with a sliding window strategy by Eq. \ref{eq:divide}}
                \STATE {Stack the patches sequentially into a volume along the depth dimension by Eq. \ref{eq:stack}}
                \STATE {Reshape the volume to match the original image's height and width by Eq. \ref{eq:reshape}}
                \STATE {\PyComment{Forward (w/o gradient) to the teacher model $f(\cdot, \theta^\prime)$}}
                \STATE {$\bar{Y}^v_i = f(X^v_i, \theta^\prime)$ and $\bar{Y}^u_j = f(X^u_j, \theta^\prime)$}
                \STATE {\PyComment{Perform Frequency and Spatial Mix-Up}}
                \STATE {Perform Frequency Mix-Up by Eq. \ref{eq:fmix} to obtain $\tilde{X}^v_{i}$ and $\tilde{X}^u_{j}$}
                \STATE {Perform Spatial Mix-Up by Eq. \ref{eq:smix} to obtain $\bar{\tilde{X}}^u_{j}$ and $\bar{\tilde{Y}}^u_{j}$}
                \STATE {\PyComment{Forward to the student model $f(\cdot, \theta)$}}
                \STATE {$\hat{Y}^l_i = f(X^l_i, \theta)$, $\hat{Y}^v_i = f(X^v_i, \theta)$, and $\hat{Y}_j = f(\bar{\tilde{X}}_j, \theta)$}
                \STATE {Compute the losses $\mathcal{L}_s$ and $\mathcal{L}_u$ by Eq. \ref{eq:loss_s} and Eq. \ref{eq:loss_u}}
                \STATE {Update $f(\cdot; \theta)$ using optimizer}
                \STATE {Update $f(\cdot; \theta^\prime)$ using EMA}
            \ENDFOR
        \STATE \textbf{return} $f(\cdot; \theta)$
    \end{algorithmic}
\end{algorithm}
%---------------------------------------------------------------

\section{Method}
In the setting of barely-supervised medical image segmentation (BSS), 
the training set $\mathcal{D} = \{\mathcal{D}^L, \mathcal{D}^U\}$ includes a barely labeled set $\mathcal{D}^L = \{(X^l_i, y^l_{i(k)})_{i=1}^{N_L}\}$ and an unlabeled set $\mathcal{D}^U = \{(X^u_j)_{j=1}^{N^U}\}$, where $X^l_i$/$X^u_j$ denotes the $i_{th}/j_{th}$ labeled/unlabeled image, $y^l_{i(k)}$ is the \textbf{single-slice annotation} of the $k_{th}$ slice $x^l_{i(k)}$ for the $i_{th}$ labeled image $X^l_i$, and $N^L$ and $N^U$ ($N^U >> N^L$) are the numbers of labeled and unlabeled samples.
Note that we take both labeled and unlabeled images as the unlabeled set, i.e., $\mathcal{D}^U = \{(X^l_i)_{i=1}^{N^L}, (X^u_j)_{j=1}^{N^U}\}$.
BSS aims to train a segmentation model $f(\cdot; \theta)$ from the partially and incompletely annotated training set.

%---------------------------------------------------------------
\begin{figure*}[!t]
\includegraphics[width=\textwidth]{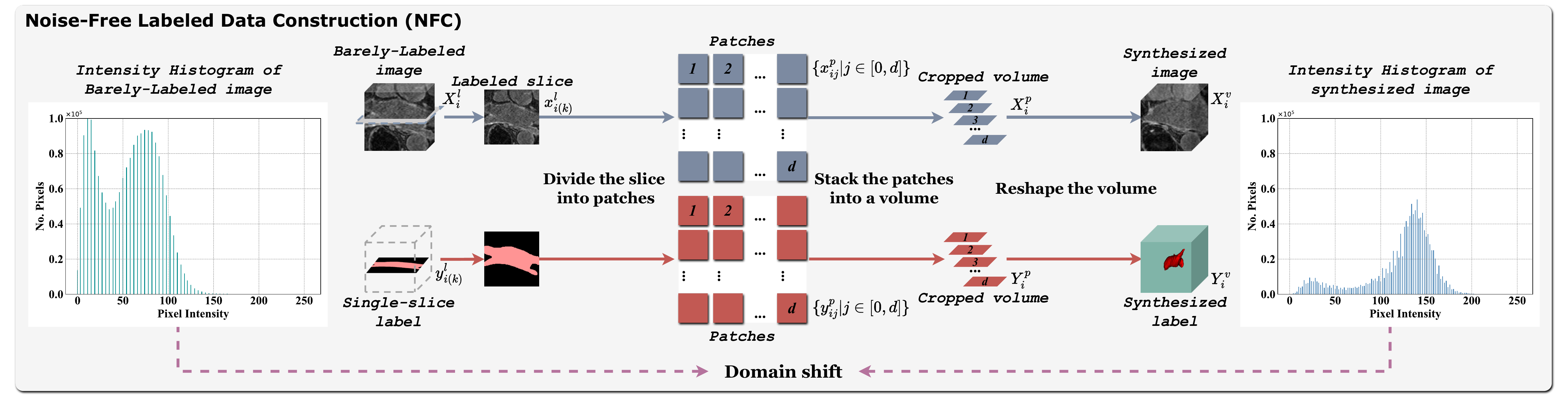}
\caption{Schematic diagram of the proposed Noise-Free Labeled Data Construction (NFC). It synthesizes volumetric image-label pairs using only barely-labeled slices, including three steps: 1) Divide the slice into patches with a sliding window strategy, 2) Stack the patches sequentially into a volume along the depth dimension, and 3) Reshape the volume to match the original image's height and width.} 
\label{fig:bva_nfc}
\end{figure*}
%---------------------------------------------------------------

%---------------------------------------------------------------
\begin{figure}[!t]
\includegraphics[width=\columnwidth]{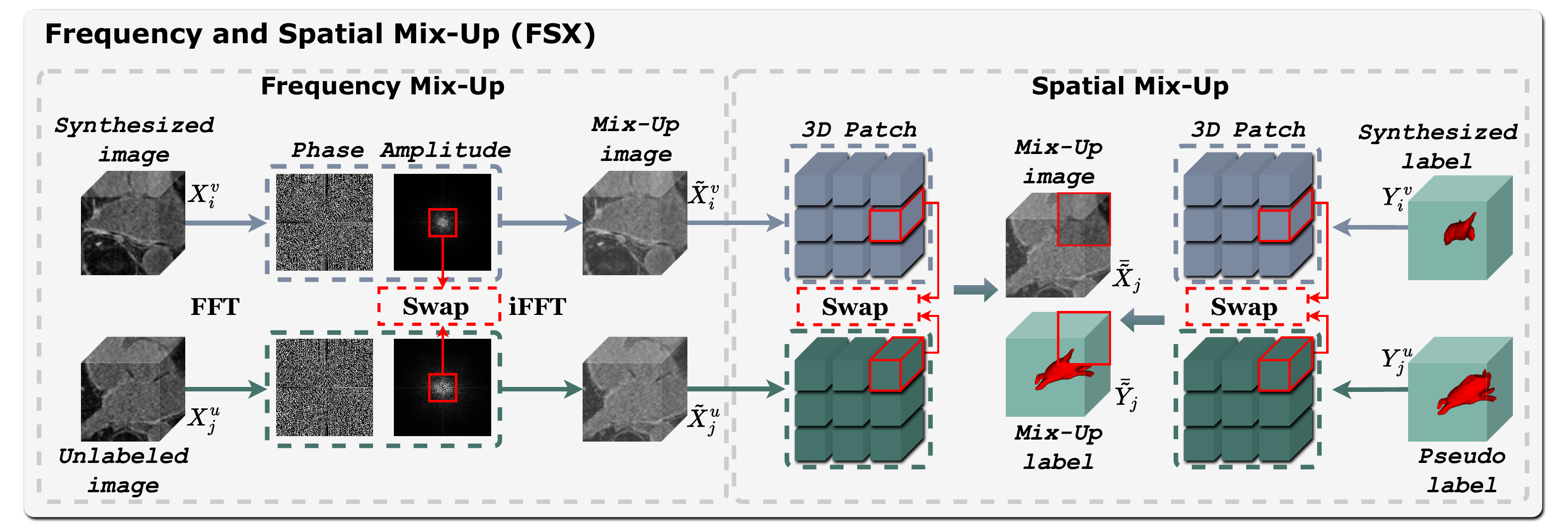}
\caption{Detailed procedure of the proposed Frequency and Spatial Mix-Up (FSX). It performs frequency and spatial Mix-Up to generate perturbed images. Note that since the spatial Mix-Up operation is not semantic-preserving, it should be conducted on images and segmentation maps simultaneously.} 
\label{fig:bva_fsx}
\end{figure}
%---------------------------------------------------------------

\subsection{Barely-Supervised Learning via Unsupervised Domain Adaptation (BvA)}
Given the limitations of the registration-based paradigm, we explore a new solution for BSS: training a segmentation model using synthesized volumetric image-label pairs generated solely from barely annotated slices as the labeled set.
Since the statistics of a single slice cannot represent those of a volumetric image, instead of solving BSS in an SSL scheme as did in previous registration-based methods \cite{li2022pln,cai2023orthogonal}, we propose a novel perspective to formulate BSS as a UDA problem [Source domain: synthesized images; Target domain: original images].
To realize this idea, we introduce a novel BSS framework, named \textbf{B}arely-supervised learning \textbf{v}ia unsupervised domain \textbf{A}daptation (BvA).
As illustrated in Fig. \ref{fig:bva_framework}, 
BvA includes two major components: 1) a noise-free labeled data construction algorithm (NFC) for constructing a complete volumetric labeled set from barely annotated data [Fig. \ref{fig:bva_nfc}] and 2) a frequency and spatial Mix-Up strategy (FSX) to mitigate domain shifts between the synthesized images and the original images [Fig. \ref{fig:bva_fsx}].
In the training phase, BvA builds upon a mean-teacher paradigm \cite{tarvainen2017mean} to leverage both the synthesized labeled data and unlabeled images, where the parameters $\theta^\prime$ of the teacher model $f(\cdot, \theta^\prime)$ is updated by an exponential moving average (EMA) of the student model  $f(\cdot, \theta)$'s parameters $\theta$ in each iteration $t$: $\theta^\prime_t = \alpha \theta^\prime_{t-1} + (1-\alpha) \theta_t$. 
The detailed training procedure of BvA is shown in Algorithm \ref{alg:BvA}.
In the testing stage, BvA only requires a single segmentation model, i.e., the student model.

\subsubsection{Noise-Free Labeled Data Construction (NFC)}
\label{subsubsec:nfc}
The key challenge in BSS is how to generate volumetric image-label pairs from barely-labeled images for constructing a complete volumetric labeled set.
Inspired by the observation that the inter-patch similarity in a slice resembles the inter-slice continuity of a volume, we develop NFC [Fig. \ref{fig:bva_nfc}] to synthesize volumetric image-label pairs using only single-annotated slices.
Let $\mathrm{Divide}(\cdot)$, $\mathrm{Stack}(\cdot)$, and $\mathrm{Resize}(\cdot)$ denote the dividing, stacking, and reshaping functions, respectively. NFC involves the following steps\footnote{Note that these operations are applied to both the single-annotated slice $x^l_{i(k)}$ its corresponding single-slice label $y^l_{i(k)}$. For illustration brevity, we have omitted the equations for $y^l_{i(k)}$, as the operations applied to both $x^l_{i(k)}$ and $y^l_{i(k)}$ are identical.}:

\noindent
\textbf{1) Divide} a single-annotated slice $x^l_{i(k)}$ (and its corresponding single-slice label $y^l_{i(k)}$) into patches using a slide window strategy: 
\begin{equation}
    \{x^p_{ij} \in \mathbb{R}^{H_p \times W_p} \,|\, j\in[1,d]\} = \mathrm{Divide}(x^l_{i(k)}, [k, s])
    \label{eq:divide}
\end{equation}
Given the slice $x^l_{i(k)} \in \mathbb{R}^{H_s \times W_s}$ and the sliding window with a window size $k$ and stride $s$, the $\mathrm{Divide}(\cdot)$ operation results in: $H_p = k$, $W_p = k$, and $d = (\frac{H_s - k}{s} + 1) \times (\frac{W_s - k}{s} + 1)$, where $d$ denotes the total number of divided patches per slice.

\noindent
\textbf{2) Stack} the patches $\{x^p_{ij}|j\in[1,d]\}$ sequentially into a cropped volume along the depth dimension: 
\begin{equation}
    X^p_i = \mathrm{Stack}(\{x^p_{ij}|j\in[1,d]\})
    \label{eq:stack}
\end{equation}

\noindent
\textbf{3) Reshape} the cropped volume $X^p_i$ to the same height and width as the original volumetric image $X^l_i$:
\begin{equation}
    X^v_i = \mathrm{Resize}(X^p_i, [H_s, W_s])
\label{eq:reshape}
\end{equation}

We determine $s$ and $k$ according to the following criteria: choosing a larger $k$ to guarantee the similarity between patches and the corresponding slices while setting an appropriate $s$ to ensure that the divided patches are sufficient for constructing images with the number of slices similar to the original images (Please refers to Section \ref{subsubsec:hyperparameters} for the detailed investigation of these hyperparameters.).

\subsubsection{Frequency and Spatial Mix-Up (FSX)}
We assume that a well-generalized model should behave smoothly across both source and target domains under small perturbations in both the style and content (shape) of images. Based on this assumption, FSX [Fig. \ref{fig:bva_fsx}] applies frequency and spatial Mix-Up perturbations to bridge the style and content gaps between the synthesized and original images, respectively.

Specifically, FSX enforces the model's predictions to remain invariant under frequency Mix-Up, which perturbs the style while preserving the content information of images \cite{yang2020fda,xu2021fourier}. 
Meanwhile, the model's prediction under a spatial Mix-Up that perturbs content by mixing image regions is regularized to be consistent with the prediction for a perturbed image under the same spatial Mix-Up. 
The above-mentioned procedure can be formulated as follows.

\noindent
\textbf{1) Frequency Mix-Up (FX)} performs Mix-Up between the amplitude components of an original image $X^u_{i}$ and a synthesized image $X^v_{i}$:
\begin{equation}
\mathcal{A}(X) = (1-\alpha)(1-\Omega)\mathcal{A}(X^u_{j}) + \alpha\Omega\mathcal{A}(X^v_{i})
\label{eq:fmix}
\end{equation}
where $\mathcal{A}(\cdot)$ denotes the mixed amplitude component, $\alpha$ controls the Mix-Up strength, and $\Omega$ is a center rectangle binary mask used to determine the Mix-Up range of amplitude spectrum \cite{yang2020fda,xu2021fourier}.
Then, the style-perturbed images are generated by inverse Fourier transformation on the mixed amplitude component and the original phase components: $\tilde{X}^v_i = \mathcal{F}^{-1}[\mathcal{A}(X)\exp{(j\mathcal{P}(X^v_i))}]$.

\noindent
\textbf{2) Spatial Mix-Up (SX)} involves the CutMix \cite{yun2019cutmix}
operation between the original images and the synthesized images:
\begin{equation}
    \bar{X}_{j} = X^u_{j} \times M + X^v_{i} \times (1 - M)
\label{eq:smix}
\end{equation}
where $M$ is a random binary mask for image region mixing.
Correspondingly, the CutMix operation should be applied to the segmentation maps:
$\bar{Y}_{i} = \bar{Y}^u_{j} \times M + \bar{Y}^v_{i} \times (1 - M)$,
where $\bar{Y}^u_{i}$ and $\bar{Y}^p_{i}$ are the pseudo labels of $X^u_{i}$ and $X^v_{i}$.

%---------------------------------------------------------------
\begin{table*}[!t]
\centering
\caption{Comparison with SOTA methods on the LA dataset with 5\% and 10\% labeled data. The best results are highlighted in \textbf{bold}.}
% \resizebox{\textwidth}{!}{
\begin{tabular}{clcccc|cccc}
\toprule[1pt]
\multicolumn{2}{c}{\multirow{3}{*}{Method}} & \multicolumn{4}{c|}{Barely-Supervised Segmentation}  & \multicolumn{4}{c}{Semi-Supervised Segmentation} \\   \cline{3-10}  & & \multicolumn{2}{c|}{5\%} & \multicolumn{2}{c|}{10\%} & \multicolumn{2}{c|}{5\%} & \multicolumn{2}{c}{10\%} \\ \cline{3-10}
 & & DSC (\%) $\uparrow$  & ASD $\downarrow$  & DSC (\%) $\uparrow$  & ASD $\downarrow$  & DSC (\%) $\uparrow$  & ASD $\downarrow$  & DSC (\%) $\uparrow$  & ASD $\downarrow$ \\ \midrule[1pt]
% \multirow{3}{*}{SSL\quad }  
& UA-MT \cite{yu2019uncertainty}  & 65.04  & 8.85  & 72.57  & 7.32  & 75.38   & 4.23    & 88.47   &    2.49 \\  
& CPS \cite{chen2021semi}  & 70.51  & 7.62 & 70.80  & 12.51  & 87.23 & 2.49  & 89.81 & 1.73 \\ 
& FixMatch \cite{sohn2020fixmatch} & 67.67  & 8.39  & 72.51  & 6.93  & 85.80  & 3.39  & 90.00  & 1.66 \\  
& UniMatch \cite{yang2023revisiting}  & 72.61  & 7.70 & 76.43  & 5.64   & 88.73  & 2.72  & 89.82  & 2.36 \\ \midrule \midrule
% \multirow{3}{*}{BSL\quad }  
& PLN \cite{li2022pln}   & 66.46  & 13.34  & 75.48  & 7.66  & \multicolumn{4}{c}{/}\\
& DeSCO \cite{cai2023orthogonal}  & 66.84  & 14.03  & 76.21  & 6.60   & \multicolumn{4}{c}{/}\\
& SPSS \cite{su2024self}  & 68.49  & 12.44  & 80.50  & 7.37  & \multicolumn{4}{c}{/}\\
& BvA (ours)  & \textbf{87.40}  & \textbf{2.37}  & \textbf{88.81}  & \textbf{1.76}  & \textbf{90.72}  & \textbf{1.58}  & \textbf{91.49}  & \textbf{1.40} \\ 
\bottomrule[1pt]
\end{tabular}
% }
\label{Tab:la}
\end{table*}
%---------------------------------------------------------------

\subsubsection{Training Objective}
The training loss of BvA is defined as:
\begin{equation}
    \mathcal{L} = \mathcal{L}_{s} + \mathcal{L}_{u}
    \label{eq:loss}
\end{equation}
where $\mathcal{L}_{s}$ and $\mathcal{L}_{u}$ represent the supervised and unsupervised losses, respectively.
Concretely, $\mathcal{L}_{s}$ includes two terms for the barely-annotated data and the 
constructed complete labeled set respectively: 
\begin{equation}
    \mathcal{L}_{s} = \mathcal{L}_{seg}\left(f\left(X^l_{i};\theta\right), y^l_{i(k)}\right) + \mathcal{L}_{seg}\left(f\left(X^v_i;\theta\right), Y^v_i\right)
    \label{eq:loss_s}
\end{equation}
where $\mathcal{L}_{seg}$ denotes a segmentation criterion. 
Moreover, $\mathcal{L}_{u}$ involves consistency regularization between the predicted segmentation maps for the original and perturbed images: 
\begin{equation}
    \mathcal{L}_{u} = \mathcal{L}_{seg}\left(f\left(\bar{\tilde{X}}_j;\theta\right), \bar{\tilde{Y}}_j\right)
    \label{eq:loss_u}
\end{equation}
where $\bar{\tilde{X}}_j$ denotes the perturbed image and $\bar{\tilde{Y}}_j$ refers to the mixed pseudo label.

\section{Experiments and Results}
\label{sec:experiments_and_results}
\subsection{Datasets}
We evaluate the proposed BvA on the Left Atrial Segmentation 2018 (LA) \cite{xiong2021global} and Brain Tumor Segmentation 2020 (BraTS)  \cite{menze2015multimodal,bakas2017advancing,bakas2018identifying} datasets.

\noindent
\textbf{LA} contains 100 gadolinium-enhanced MRI scans. Following \cite{yu2019uncertainty}, we split LA into 80 samples for training (where we further divided 80 samples into 70 for training and 10 for validation) and 20 samples for testing (i.e., \textit{train} : \textit{val} : \textit{test} = $7:1:2$). 

\noindent
\textbf{BraTS} consists of 369 multi-modal MRI scans with four modalities (FLAIR, T1, T1Gd, and T2). 
We divide BraTS into 258, 37, and 74 subjects (i.e., $7:1:2$) for training, validation, and testing, respectively.

\subsection{Implementation Details}

\noindent
\textbf{Experimental environment:} 
All experiments are conducted under the same environment (NVIDIA Quadro RTX 6000 GPU with 24G GPU memory; PyTorch 1.11.0, CUDA 11.3). All methods are optimized using the AdamW optimizer \cite{kingma2014adam} with a constant learning rate of $1e-4$ for 500 epochs.

\noindent
\textbf{Framework:}
Following Mean-Teacher \cite{tarvainen2017mean}, the EMA decay $\alpha$ is set to 0.99.
We employ V-Net \cite{milletari2016v} as the fully-supervised segmentation backbone.
Dice loss is used as the segmentation criterion $\mathcal{L}_{seg}$.

\noindent
\textbf{Data:}
In the training phase, we randomly crop $80 \times 112 \times 112 \, (Depth \times Height \times Width)$ patches for LA and BraTS. 
We set the window size $k$ to half the size of the original slice and the stride $s = 8$ (Detailed analysis for these hyperparameters is provided in Section \ref{subsubsec:hyperparameters}). 

\noindent
\textbf{Evaluation metrics:} 
Dice similarity coefficient (DSC) and average surface distance (ASD) are employed to evaluate segmentation performance in the experiments.

\subsection{Comparison with SOTA}
We compare BvA with SOTA methods under both barely-supervised and semi-supervised segmentation settings on the LA and BraTS datasets. 
The compared methods include: SSL (UA-MT \cite{yu2019uncertainty}, CPS \cite{chen2021semi}, FixMatch \cite{sohn2020fixmatch}, and UniMatch \cite{yang2023revisiting}), BSL (PLN \cite{li2022pln}, DeSCO \cite{cai2023orthogonal}, and SPSS \cite{su2024self}).
The labeled data are set as 5\% and 10\% of the entire training set respectively, with single-slice annotations for barely-supervised segmentation and complete volumetric 
annotations for semi-supervised segmentation. 
Note that for registration-based BSS methods, we only report their results in the BSL setting, as the registration module becomes non-functional in the SSL setting, causing these methods to degrade to a vanilla mean-teacher framework.

%---------------------------------------------------------------
\begin{table*}[!t]
\centering
\caption{Comparison with SOTA methods on the BraTS dataset with 5\% and 10\% labeled data. The best results are highlighted in \textbf{bold}.}
% \resizebox{\textwidth}{!}{
\begin{tabular}{clcccc|cccc}
\toprule[1pt]
\multicolumn{2}{c}{\multirow{3}{*}{Method}} & \multicolumn{4}{c|}{Barely-Supervised Segmentation}  & \multicolumn{4}{c}{Semi-Supervised Segmentation} \\   \cline{3-10} 
& & \multicolumn{2}{c|}{5\%} & \multicolumn{2}{c|}{10\%} & \multicolumn{2}{c|}{5\%} & \multicolumn{2}{c}{10\%} \\ \cline{3-10}
& & DSC (\%) $\uparrow$  & ASD $\downarrow$  & DSC (\%) $\uparrow$  & ASD $\downarrow$  & DSC (\%) $\uparrow$  & ASD $\downarrow$  & DSC (\%) $\uparrow$  & ASD $\downarrow$ \\ \midrule[1pt]
 % \multirow{3}{*}{SSL\quad } 
& UA-MT \cite{yu2019uncertainty}  & 19.81  & 20.67  & 23.17  & 37.31  & 43.62   & 2.66    & 52.89   &    2.94 \\  
& CPS \cite{chen2021semi}  & 13.82  & 50.64  & 34.07  & 32.81  & 65.20  & 1.81  & 66.52  & 2.25\\
& FixMatch \cite{sohn2020fixmatch} & 53.94  & 5.13  & 56.29  & 2.86  & 58.65  & 3.13  & 63.08  & 2.14 \\
& UniMatch \cite{yang2023revisiting}  & 51.44  & 9.85  & 56.06  & 4.52  & 60.29  & 2.34  & 63.96  & \textbf{1.76} \\ \midrule \midrule
 % \multirow{3}{*}{BSL\quad } 
& PLN \cite{li2022pln}    & 5.41  & 54.09  & 7.44  & 57.84  & \multicolumn{4}{c}{/}\\
& DeSCO \cite{cai2023orthogonal}    & 5.15  & 57.16  & 7.40  & 54.18  & \multicolumn{4}{c}{/}\\
& SPSS \cite{su2024self}  &  20.89  &  38.37  &  17.26  & 40.71  & \multicolumn{4}{c}{/} \\
& BvA (ours)  & \textbf{58.42}  & \textbf{3.83}  & \textbf{60.47}  & \textbf{3.18}  & \textbf{66.64} & \textbf{1.75}  & \textbf{67.76}  & 1.79\\
\bottomrule[1pt]
\end{tabular}
% }
\label{Tab:brats}
\end{table*}
%---------------------------------------------------------------

%---------------------------------------------------------------
\begin{figure*}[!t]
\centering
\includegraphics[width=\textwidth]{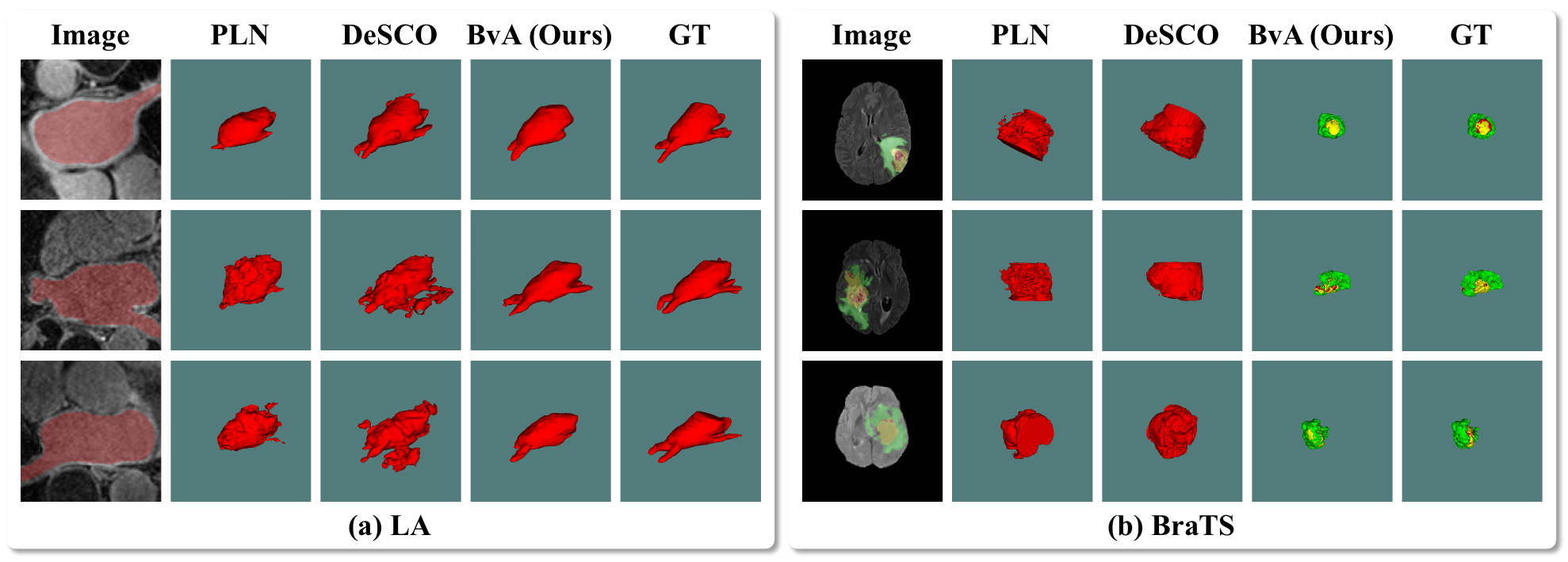} 
\caption{Qualitative examples on the LA and BraTS datasets. 
GT: Ground Truth.}
\label{fig:qualitative}
\end{figure*}
%---------------------------------------------------------------

\subsubsection{Results on LA} 
As reported in Table \ref{Tab:la}, 
BvA sets a new state-of-the-art with 87.40\% and 88.81 in terms of DSC for barely-supervised segmentation on the LA dataset under 5\% and 10\% barely-annotated labeled data. 
For instance, BvA consistently outperforms the registration-based BSS methods, i.e., PLN \cite{li2022pln}, DeSCO \cite{cai2023orthogonal}, and SPSS \cite{su2024self} by a large margin.
The inferior performance of the registration-based methods implies that the noisy pseudo labels generated by image registration degrade the training process of these models.
Besides, the SSL methods yield unsatisfactory results in the BSS setting, which can be attributed to the failure to extract volumetric shape information due to the lack of complete volumetric annotations. 
In semi-supervised segmentation, our BvA also achieves the best performance among the compared methods. These results suggest the versatility of BvA for both SSL and BSS scenarios.

\subsubsection{Results on BraTS}
Tumor segmentation is more challenging than organ segmentation due to the heterogeneity of tumors.
Table \ref{Tab:brats} shows the averaged performance of brain tumor segmentation (three classes: enhancing tumor, peritumoral edema, and necrotic tumor core) on the BraTS dataset.
On the one hand, BvA achieves 58.42\% and 60.47\% DSC, and 3.83 and 3.18 ASD, respectively, under 5\% and 10\% barely-labeled data, presenting considerable improvements compared with other methods.
As image registration cannot capture the heterogeneity of tumors, PLN \cite{li2022pln}, DeSCO \cite{cai2023orthogonal}, and SPSS fail to delineate brain tumors accurately, obtaining unsatisfactory results.
Besides, without the impact of registration noise, the SSL methods obtain relatively higher performance than the registration-based approaches.
On the other hand, compared with the semi-supervised state-of-the-art, UniMatch \cite{yang2023revisiting}, BvA achieves significant gains of 6.36\% and 3.80\% in terms of DSC.
These results further demonstrate the superiority of BvA over the state-of-the-art in both the barely-supervised and semi-supervised medical image segmentation.

\subsection{Qualitative Results}
In Fig. \ref{fig:qualitative}, we present segmentation examples from the LA and BraTS datasets under 5\% barely labeled data. The proposed BvA demonstrates superior qualitative results for both organ and tumor segmentation compared to registration-based methods, i.e., PLN \cite{li2022pln} and DeSCO \cite{cai2023orthogonal}. This phenomenon can be attributed to the detrimental impact of registration noise, which degrades or even overwhelms the training processes of PLN and DeSCO, leading to inferior segmentation outcomes. These results are consistent with the performance metrics reported in Table \ref{Tab:la} and Table \ref{Tab:brats}, further validating the superiority of BvA for barely-supervised volumetric medical image segmentation.

%---------------------------------------------------------------
\begin{table*}[!t]
\centering
\caption{Ablation study of the proposed BvA on the LA dataset under 5\% labeled data. F/SX: Frequency/Spatial Mix-Up.
The best results are highlighted in \textbf{blod}.}
% \resizebox{\columnwidth}{!}{
\begin{tabular}{c|cccc|cc}
\toprule[1pt]
\multirow{2}{*}{Method} & \multicolumn{4}{c|}{Component} & \multicolumn{2}{c}{5\%}\\
& \quad MT \quad & \quad NFC \quad & \quad FX \quad & \quad SX \quad & DSC (\%) $\uparrow$ & ASD $\downarrow$ \\
\midrule[1pt]
Baseline	&  $\surd$ &  &  &  & 67.06  & 12.74 \\
Baseline + NFC   &  $\surd$ & $\surd$ &  &  & 72.96  & 6.17 \\
Baseline + NFC + FX  &  $\surd$ & $\surd$ & $\surd$ &  & 85.49  & 3.21 \\
Baseline + NFC + SX  &  $\surd$ & $\surd$ &  & $\surd$ & 79.53  & 3.65 \\
Baseline + NFC + FSX (BvA)  &  $\surd$ & $\surd$ & $\surd$ & $\surd$ & \textbf{87.40}  & \textbf{2.37} \\
\bottomrule[1pt]
\end{tabular}
\label{Tab:ablation}
\end{table*}
%---------------------------------------------------------------

%---------------------------------------------------------------
\begin{figure*}[!t]
\centering
\includegraphics[width=0.85\textwidth]{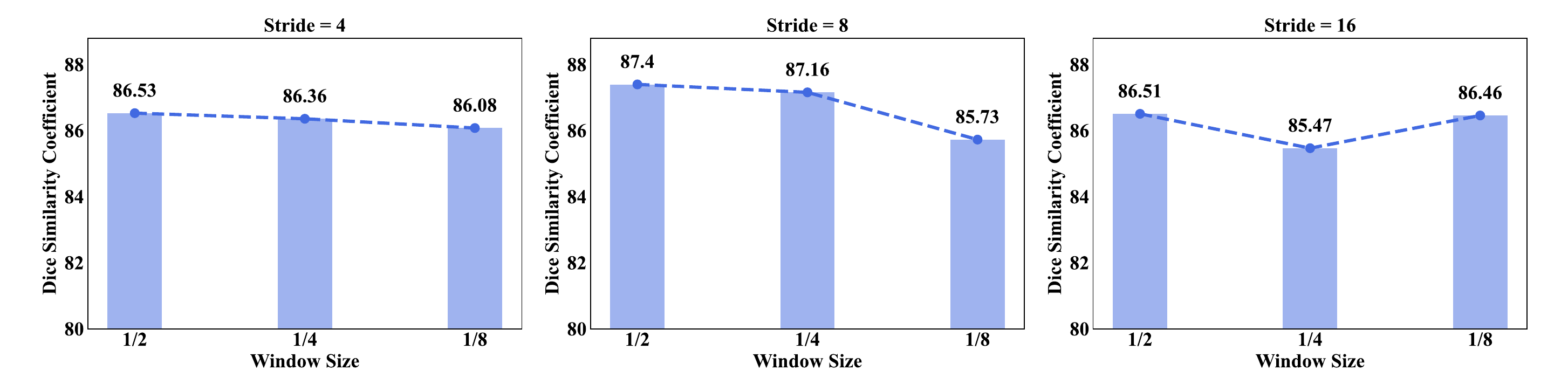} 
\caption{Investigation of the two hyperparameters in the proposed noisy-free labeled data construction algorithm (NFC): \textbf{the window size $k$} and \textbf{the stride $s$}. Note that $k$ is set to a proportion of the size of the original slice.}
\label{fig:hyperparameters}
\end{figure*}
%---------------------------------------------------------------

\subsection{Ablation Study}
\label{subsec:ablation}

\subsubsection{Effectiveness of Each Component}
\label{subsubsec:effectiveness_of_eachcomponent}
We conduct an ablation study on the LA dataset under 5\% barely-annotated labeled data to investigate the effectiveness of the components of BvA.
In Table \ref{Tab:ablation}, one can observe that the segmentation performance gradually increases as each component is introduced into our method. Specifically, with NFC to construct a complete volumetric labeled set from the barely annotated labeled set, the DSC score increases from 67.06\% to 72.96\% and the ASD value improves from 12.74 to 6.17. Then, FX and SX are adopted to address the style and content domain shifts between the synthesized and the original images, further bringing performance improvements of 12.53\% and 6.57\% in terms of DSC, respectively. Finally, when the domain shifts are alleviated through both frequency and spatial domain perturbations, BvA improves the segmentation performance to a DSC of 87.40\% and an ASD of 2.37. In light of the above, the performance improvement demonstrates that each component contributes positively to the proposed BvA.

\subsubsection{Investigation of hyperparameters}
\label{subsubsec:hyperparameters}
As depicted in Fig. \ref{fig:hyperparameters}, we conducted an experiment on the LA dataset with 5\% barely-annotated labeled data to investigate the sensitivity of NFC on the window size $k$ and stride $s$. 
Note that $k$ is set to a proportion of the size of the original slice. 
The results are measured using DSC. 
It can be observed that the setting where the window size $k = 1/4$ and stride $s = 8$ leads to the best DSC score of 87.40\%. Reducing the window size leads to performance decreases, and both smaller and larger strides also result in performance drops.
The result accords with our determining criterion mentioned in Section \ref{subsubsec:nfc}: a larger window size can guarantee the similarity between patches and the corresponding slices, while an appropriate stride can ensure that the divided patches are sufficient for constructing images with the length of \textit{depth} dimension similar to the original images.
Based on this experiment, we set the window size $k$ to 1/2 of the size of the original slice and the stride $s = 8$ in NFC.

%---------------------------------------------------------------
\begin{figure}[!t]
\centering
\includegraphics[width=\columnwidth]{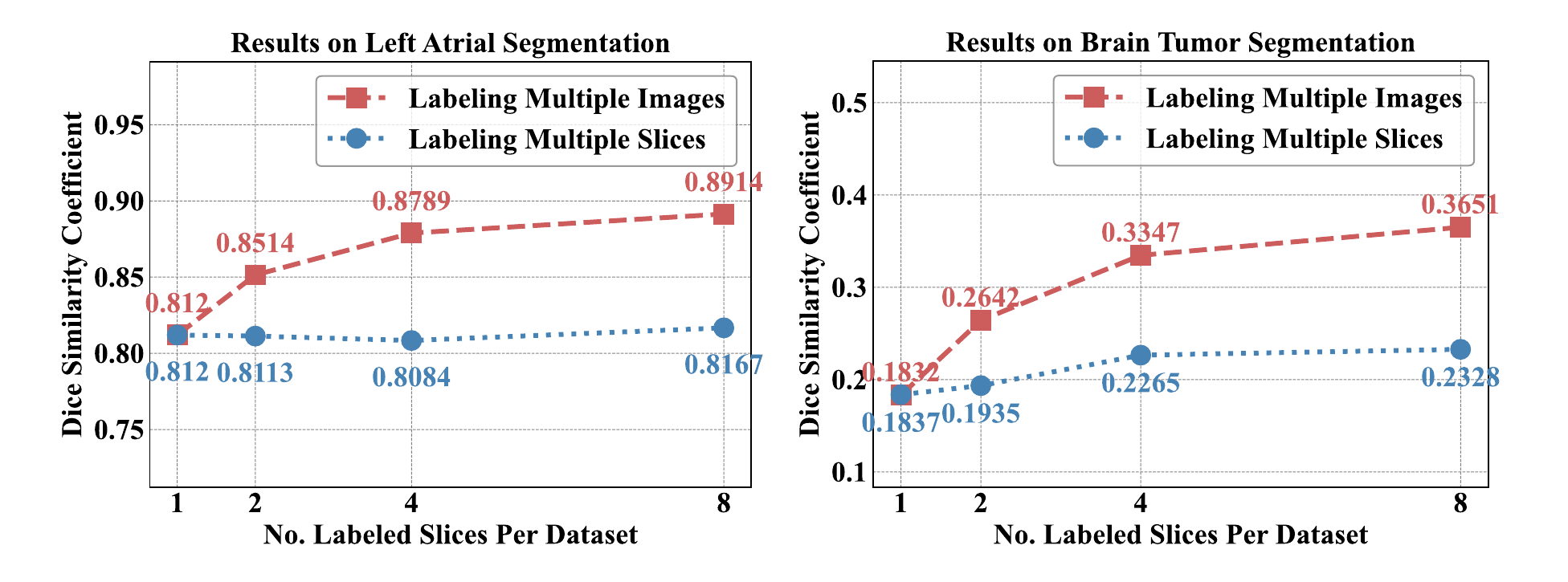} 
\caption{Analysis of different annotation configurations: different numbers of barely-annotated images per dataset (Multi-Labeled Images) vs. different numbers of labeled slices in one barely-annotated image (Mult-Labeled Slices).}
\label{fig:analysis}
\end{figure}
%---------------------------------------------------------------

\subsection{Analysis of Different Number of Labeled Slices}
We conducted an experiment on the LA and BraTS datasets to investigate the influence of labeling different number slices (1, 2, 4, 8) per dataset.
We considered two annotation strategies: 1) annotating multiple images with single-slice annotations, and 2) annotating only one image with multi-slice annotations. 
As shown in Fig. \ref{fig:analysis}, in general, the performance gradually improves as the number of labeled slices increases in both situations, since more labeled data can provide more ground truth supervision signals.
It can be observed that adopting the strategy of annotating multiple images with single-slice annotations leads to a significant improvement in segmentation performance as the number of labeled slices increases.
In the case of annotating multiple slices within a single image, the performance improvement in the left atrium segmentation task is not significant due to the high redundancy of left atrium shape information between slices; in contrast, in the brain tumor segmentation task, the heterogeneity of gliomas results in lower redundancy of tumor shape information between slices, thereby leading to a more noticeable performance improvement when annotating multiple slices within a single image.
More importantly, compared with annotating multiple slices in a single case, labeling multiple images with single-slice annotations leads to a more significant performance improvement. 
This result is reasonable, as the former situation contains more redundant supervision information due to inter-slice similarity, while the latter provides more diverse ground truth supervision signals for model training.
From this result, it can be concluded that for volumetric medical image segmentation tasks, annotating multiple images with single-slice annotations is a more effective sparse labeling strategy.

\subsection{Analysis of Different Stacking Strategies}
We further investigate the impact of different stacking strategies used in NFC on the construction of the volumetric labeled set. 
As shown in Table \ref{tab:stack_strategy}, the stacking strategies include: 1) \textit{Sequential stack}: Stacking the patches sequentially into a volume along the depth dimension, 2) \textit{Random stack}: Stacking the patches randomly into a volume along the depth dimension, and 3) \textit{Stack with noise}: Stacking the patches sequentially with random insertion of noise patches.
This experiment was conducted on the LA dataset with 5\% and 10\% labeled data. 
It can be observed that employing the \textit{Sequential stacking} strategy results in a significant performance improvement, in terms of both DSC and ASD, compared with the \textit{Random stacking} and \textit{Stacking with noise} strategies.
This is because \textit{Sequential stacking} can reserve the shape information of volumetric images, while  \textit{Random stacking} and \textit{Stacking with noise} disrupt the continuity between slices.
This phenomenon also aligns with the idea of NFC, where NFC reserves the inter-slice continuity leveraging the inter-patch similarity.

%---------------------------------------------------------------
\begin{table}[!t]
\caption{Analysis of different stacking strategies (\textit{Sequential stack}, \textit{Random stack}, and \textit{Stack with noise}) on the LA dataset with 5\% and 10\% labeled data. The best results are highlighted in \textbf{blod}.} 
\centering
\includegraphics[width=\columnwidth]{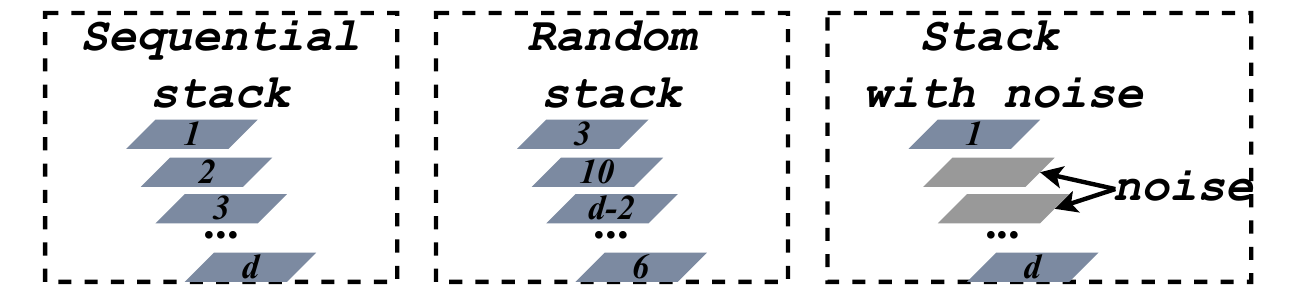}
\resizebox{\columnwidth}{!}{
\begin{tabular}{c|cccc}
\toprule[1pt]
\multirow{2}{*}{Stack strategy} & \multicolumn{2}{c|}{5\%} & \multicolumn{2}{c}{10\%}\\ \cline{2-5} 
& DSC (\%) $\uparrow$ & ASD $\downarrow$ & DSC (\%) $\uparrow$ & ASD $\downarrow$ \\
\midrule[1pt]
\textit{Sequential stack} (ours)  & \textbf{87.40}  & \textbf{2.37} & \textbf{88.81}  & \textbf{1.76} \\
\textit{Random stack}  & 83.63  & 6.78 & 85.28  & 6.23 \\
\textit{Stack with noise}  & 81.35  & 7.45 & 84.22  & 6.39 \\
\bottomrule[1pt]
\end{tabular}
}
\label{tab:stack_strategy}
\end{table}
%---------------------------------------------------------------

\section{Conclusion}
This paper initially frames barely supervised segmentation as an unsupervised domain adaptation problem, wherein we introduce a novel method, named BvA.
Our main ideas lie in the observation that inter-patch similarity in a slice resembles inter-slice continuity in a volume, as well as the assumption that a well-generalized model should exhibit smoothness across domains under small perturbations.
The experimental results on the LA and BraTS datasets under both barely-supervised and semi-supervised settings, demonstrate the effectiveness and superiority of BvA over the state-of-the-art. The results also suggest that annotating multiple images with single-slice annotations is a feasible sparse labeling strategy for volumetric medical image segmentation and is more effective than annotating a single image with multi-slice annotations.

\bibliographystyle{IEEEtran}
\bibliography{main}

\end{document}